\pdfoutput=1

\documentclass[11pt]{article}
\usepackage{authblk}
\usepackage{acl}
\usepackage{times}  
\usepackage{helvet}  
\usepackage{courier}  
\usepackage{graphicx}
\usepackage{natbib}  
\usepackage{caption}
\usepackage{algorithm}
\usepackage{algorithmic}
\usepackage{amsfonts}
\usepackage{amsmath}
\usepackage{array}
\usepackage{color}
\usepackage{times}
\usepackage{latexsym}
\usepackage{newfloat}
\usepackage{pifont}
\usepackage{listings}
\usepackage{float}
\usepackage{multirow}
\newcommand{\xmark}{\ding{55}}%
\usepackage[T1]{fontenc}

\usepackage[utf8]{inputenc}
\usepackage{blindtext}
\usepackage{microtype}
\usepackage[misc]{ifsym}
\title{MWP-BERT: Numeracy-Augmented Pre-training for Math Word Problem Solving}

\author[1]{\textbf{Zhenwen Liang}}
\author[2]{\textbf{Jipeng Zhang}}
\author[3]{\textbf{Lei Wang}}
\author[4]{\textbf{Wei Qin}}
\author[5]{\textbf{Yunshi Lan}}
\author[6]{\textbf{Jie Shao}}
\author[1]{\textbf{Xiangliang Zhang}\textsuperscript{\tiny \Letter}}
\affil[1]{University of Notre Dame, \texttt{\;\{zliang6, xzhang33\}@nd.edu}}
\affil[2]{Hong Kong University of Science and Technology, \texttt {jzhanggr@conect.ust.hk}}
\affil[3]{Singapore Management University, \texttt{\;lei.wang.2019@phdcs.smu.edu.sg}}
\affil[4]{Hefei University of Technology,\texttt{\;qinwei.hfut@gmail.com}}
\affil[5]{East China Normal University, \texttt{\; yslan@dase.ecnu.edu.cn}}
\affil[6]{University of Electronic Science and Technology of China, \texttt{\;shaojie@uestc.edu.cn}}

\begin{document}

\maketitle

\begin{abstract}

Math word problem (MWP) solving faces a dilemma in number representation learning. In order to avoid the number representation issue and reduce the search space of feasible solutions, existing works striving for MWP solving usually replace real numbers with symbolic placeholders to focus on logic reasoning. However, different from common symbolic reasoning tasks like program synthesis and knowledge graph reasoning, MWP solving has extra requirements in numerical reasoning. In other words, instead of the number value itself, it is the reusable numerical property that matters more in numerical reasoning. Therefore, we argue that injecting numerical properties into symbolic placeholders with contextualized representation learning schema can
provide a way out of the dilemma in the number representation issue here. In this work, we introduce this idea to the popular pre-training language model (PLM) techniques and build MWP-BERT, an effective contextual number representation PLM. We demonstrate the effectiveness of our MWP-BERT on MWP solving and several MWP-specific understanding tasks on both English and Chinese benchmarks.

\end{abstract}

\section{Introduction}

\begin{figure}[t!]
\centering 
\includegraphics[width=0.475\textwidth]{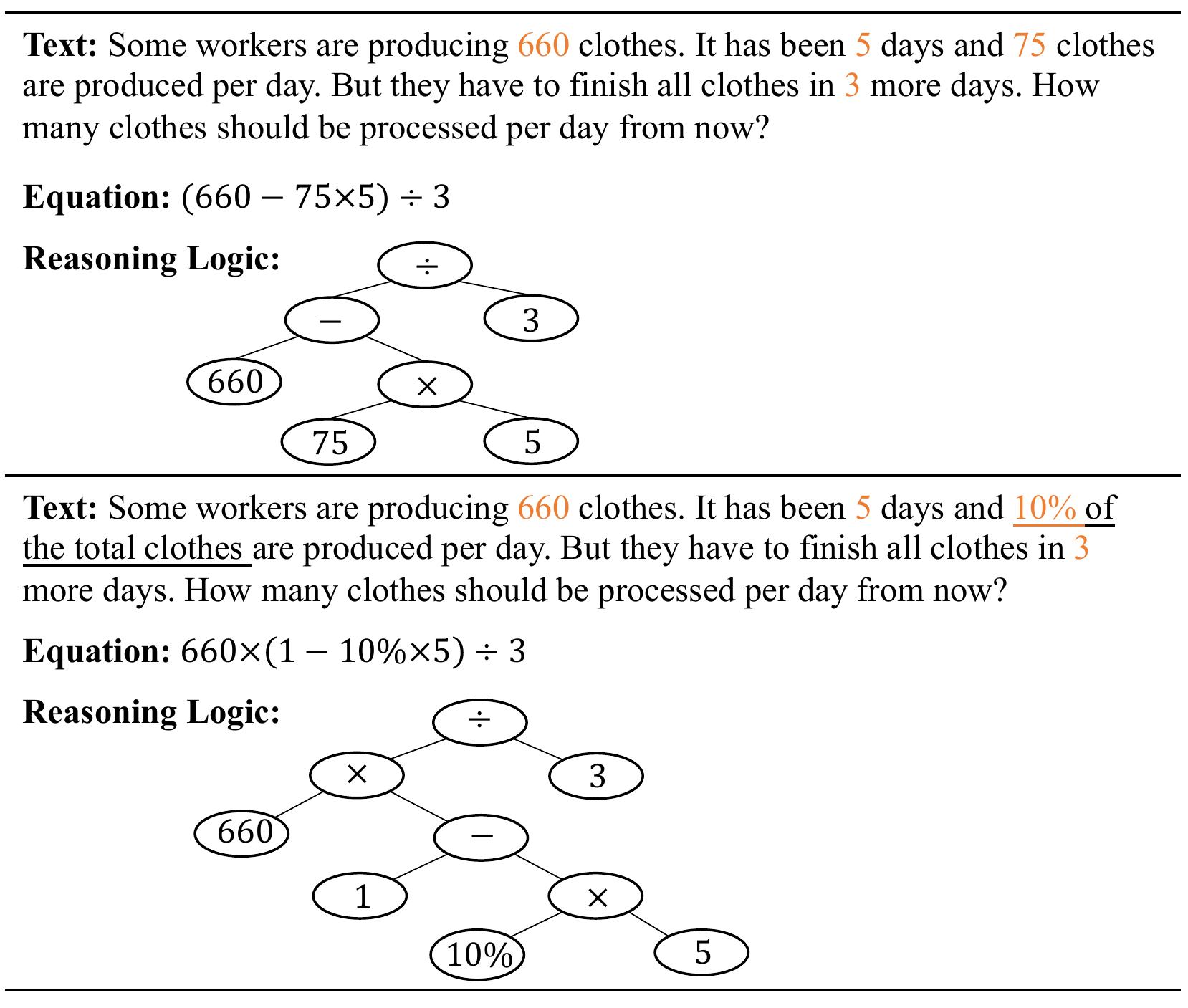}
\caption{The second question is obtained from the first one by minor modifications. However, their solution equation and corresponding equation tree structure are different from each other. This demonstrates the importance
of considering numerical value information and reasoning logic (equation tree) in contextual modeling.}
\label{fig:intro} 
\end{figure}
Recent works in math word problem (MWP) solving~\cite{wang2018translating,wang2019template,liu2019tree,li2019modeling,xie2019goal,zhang2020graph,wu2020knowledge,qin2021neural,huang2021recall,wu2021edge,yu2021improving,DBLP:conf/emnlp/ShenYLSJ0021} arrange  the  pipeline into a sequence-to-sequence framework. In brief, they use deep representation and gradient optimization as well as symbolic constraints to discover discrete symbolic combinations of operators and variants. Fundamentally, MWP solving system aims to perform symbolic reasoning by searching through a combinatorial solution space given the text description evidences. Thus, 
these neurosymbolic methods mainly focus on getting more effective semantic representations~\cite{li2019modeling,zhang2020graph,wu2020knowledge,wu2021edge,yu2021improving}, injecting symbolic constraints~\cite{wang2018translating,wang2019template,liu2019tree,xie2019goal} and how to align semantic space (text descriptions) and huge combinatorial symbolic space (symbolic solutions)~\cite{qin2021neural,DBLP:conf/emnlp/ShenYLSJ0021,huang2021recall}. This line of methods has achieved great success and is still holding the lead in various MWP solving benchmarks~\cite{wang2017deep,zhao2020ape210k,koncel2016mawps}. 

Despite the great performance achieved by the previous methods, there still exists fundamental challenges in number representation for MWP solving. More exactly, number values are required to be considered as vital evidence in solution exploration but existing works are known to be inefficient in capturing numeracy information~\cite{DBLP:conf/emnlp/WallaceWLSG19}. Intuitively, we could simply treat explicit numbers in the same way with words, i.e., assign position for all numbers in the vocabulary. However, there would be an infinite number of candidates during prediction and it would be impossible to learn their deep representations. In other words, the solution space will be extremely large and the complexity is unacceptable. Therefore, almost all existing works follow the number mapping technique~\citet{wang2017deep} to replace all numbers with symbolic placeholders (e.g., ``x1'', ``x2''). The core idea here is to get a reasonable solution space by restricting neural networks to leave out numerical characteristics and focus on logic reasoning. However, most of \textcolor{black}{the current MWP solvers}
do not consider the background knowledge in the context and are usually inefficient in capturing numeracy properties. An example is shown in Fig.~\ref{fig:intro}. Small perturbations in the problem description actually bring large variations in reasoning logic and equation. If the model simply regards ``75'' and ``10\%'' as the same placeholder ``x3'', and does not notice the small variation in the context, a wrong solution will be generated.


To this end, we incorporate several numeracy grounded pre-training objectives to inject inductive bias about numerical constraints into dynamic representations. Compared with word candidate sets, useful points in number candidate space are \textcolor{black}{scattered sparsely}. However, we identify that during prediction, what matters is the reusable numerical properties of number values. What's more, these properties do not suffer from the sparsity issues of specific values in   MWPs. Therefore, compared with assigning a prototype vector for each single number value like \cite{wu2021math}, it is more reasonable to inject the reusable numerical properties in deep representations, e.g., magnitude and number type. 
In this work, we propose to design numeracy grounded pre-training objectives to implement soft constraints between symbolic placeholders and numbers in deep representation.

\paragraph{Contributions.}
We present a suite of numeracy-augmented pre-training tasks with consideration of reasoning logic and numerical properties. More exactly, we introduce several novel pre-training tasks with access to different levels of supervision signals to make use of more available MWP data. 
\begin{itemize}
    \item One basic group of pre-training tasks is designed for the self-supervised setting. Except for masked language modeling (MLM), we give extra consideration to number-related context information by designing related objectives. 
    \item Another set of pre-training objectives is for the weakly-supervised setting that has only  answer annotations  but no equation solutions. With access to the answer value, we introduce several tasks to determine the type and the value of the answer. 
    \item The final set is for the fully-supervised setting, where both solution equation and answer are available for the MWPs. 
\end{itemize}
Besides, a group of numeracy grounded pre-training objectives is designed to leverage the corpus of MWP and encourage the contextual representation to capture numerical information. Experiments conducted on both Chinese and English benchmarks show the significant improvement of our proposed approach over all competitors. To our knowledge, this is the first approach that surpasses human performance \cite{wang2019template} in terms of MWP solving.

\section{Related Works}

\paragraph{Math Word Problems Solving.}
There exist  two major types of MWP, equation set MWP~\cite{wang2017deep,zhao2020ape210k} and arithmetic MWP~\cite{DBLP:conf/emnlp/QinLLZL20,huang2016well}. This work focuses on arithmetic MWP, which is usually paired with one unknown variable.
Along the path of the MWP solver's development, the pioneer studies use traditional rule-based methods, machine learning methods and statistical methods \cite{yuhui2010frame,kushman2014learning,shi2015automatically,koncel2015parsing}.
Afterwards, inspired by the development of sequence-to-sequence (Seq2Seq) models, MWP solving has been formulated as a neurosymbolic reasoning pipeline of translating language descriptions to mathematical equations with encoder-decoder framework \cite{wang2018translating,wang2019template,li2019modeling,zhang2020graph,yu2021improving,wu2021edge}. By fusing hard constraints into decoder~\cite{DBLP:conf/naacl/Chiang2019,liu2019tree,xie2019goal,shen2020solving,ijcai2020-555}, MWP solvers achieve much better performance then. Several works propose to utilize multi-stage frameworks~\cite{wang2019template,huang2021recall,DBLP:conf/emnlp/ShenYLSJ0021,ijcai2021} to make more robust solvers. Also, several new works made attempts to improve MWP solver beyond supervised settings~\cite{hong2021weakly,hong2021smart}. 

Among all these previous studies, the most relevant ones to our work can be categorized into two groups. First, it has been noted that number values and mathematical constraints play a significant role in supporting numerical reasoning. \citet{wu2021math} proposed several number value features to enhance encoder and \citet{qin2021neural} designed new auxiliary tasks to enhance neural MWP solvers. Compared with their work, we first introduce pre-training language model (PLM) and concentrate on representation learning to resolve numerical understanding challenges. Second, regarding the usage of  pre-training techniques for MWP solving,  \citet{DBLP:conf/emnlp/ShenYLSJ0021} introduced BART-based~\cite{DBLP:conf/acl/LewisLGGMLSZ20} MWP solver and incorporated specialized multi-task training for obtaining more effective pre-training Seq2Seq models for MWP. Compared with them, our work focuses on the number representation learning issue of MWP and achieves a more flexible pre-training representation module for MWP solving, which can be applied in various MWP related tasks other than solution generation. 

\paragraph{Numeracy-aware Pre-training Models.}
Number representation has been recognized as one of the main issues in word representation learning. Existing methods make use of value, exponent, sub-word and character methods~\cite{DBLP:conf/naacl/ThawaniPIS21} to obtain number representations for explicit number values. These methods are known to be less effective in extrapolation cases like testing with numbers not appearing in the training corpus. 

Previous related works~\cite{DBLP:conf/emnlp/AndorHLP19,DBLP:conf/emnlp/WallaceWLSG19,DBLP:conf/acl/GevaGB20} mainly focus on shallow numerical reasoning tasks shown in DROP dataset~\cite{DBLP:conf/naacl/DuaWDSS019}, which usually serves as a benchmark for evaluating numerical machine reading comprehension (Num-MRC) performance. Compared with MWP solving, Num-MRC's main focus is laid on extracting answer spans from a paragraph, which are more fault-tolerant with no needs to predict number tokens. Besides, their solution generation tasks only contain simple computations like addition/subtraction and there are only integers in DROP. More exactly, several research efforts have been made to deal with this kind of math-related reading comprehension task by synthesizing new training examples~\cite{DBLP:conf/acl/GevaGB20}, incorporating special modules considering the numerical operation~\cite{DBLP:conf/emnlp/AndorHLP19} and designing specific tokenization strategies~\cite{DBLP:conf/emnlp/0001RTER20}. Since MWP solving requires further consideration of the complex composition of reasoning logics in MWP text, the symbolic placeholder is more effective in MWP solving. Thus, instead of dealing with explicit number values, our work focuses on improving representation for symbolic placeholders by injecting numerical properties in a probabilistic way.

\begin{figure}
\centering 
\includegraphics[width=0.49\textwidth]{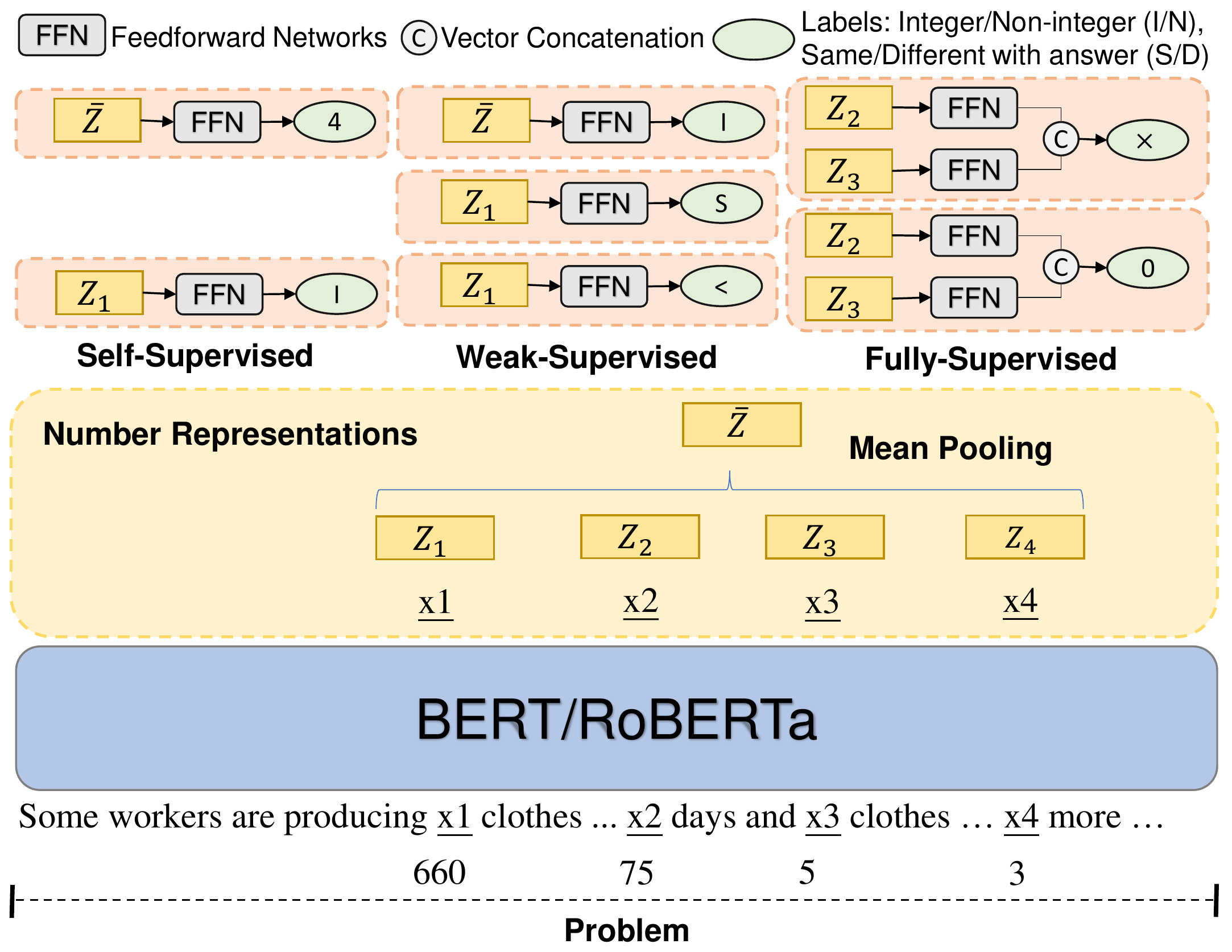} 
\caption{The overall architecture of our BERT-based MWP solver. Our method enables the solver to learn from unlabeled, incompletely labeled and fully labeled MWPs by different pre-training tasks.} 
\label{fig:network} 
\end{figure}

\section{Methodology}
\subsection{Problem Statement}
The input to an MWP solver is a textual description, we denote it as $W$ with length $m$, thus $W = \{w_1,w_2,...,w_m\}$. We also define a subset $W_q$ of $W$ which contains all the quantities appeared in $W$. The output is an equation showing how to get the final answer. We denote it as $A$ with length $n$, where $A = \{a_1,a_2,...,a_n\}$. The vocabulary of $A$ contains three parts, namely $V_{op}$, $V_{num}$ and $V_{cons}$. $V_{op}$ is the vocabulary for all operators, i.e. $+$, $-$, $\times$, $\div$ and $ ^\wedge $. The vocabulary of quantities $V_{num}$ is constructed by number mapping \cite{wang2017deep}, which transforms quantities in different MWPs into a unified representation. More specifically, $V_{num}$ does not contain the actual value of quantities appeared in $W$, and those quantities are denoted as $\{n_1,n_2,...,n_k\}$, where $n_i$ means the $i$-th number from $W$ and $k$ is the maximum number of quantities in $V_{num}$ in order to fix the size of it. $V_{cons}$ contains necessary constant values e.g.,  $\pi$.


\subsection{PLM Encoder}~\label{sec:train_obj}
Our PLM encoder maps the problem description $W$ into a representation matrix $Z \in \mathbb{R}^{m*h}$ where $h$ is the dimension of the hidden feature.

\begin{equation}
\label{encoder}
    Z = encoder(W).
\end{equation}

The representation vector corresponding to each word in $Z$ will be used in the decoding process for generating the solution. 

An overview of pre-training objectives and our model architecture is shown in Figure~\ref{fig:network}. In general, pre-training objectives are designed to inject contextual priori and numerical properties as soft constraints for representation learning. They are categorized into three types given provided training signals, i.e., self-supervised, weakly-supervised, and fully-supervised.

\subsection{Self-supervised Objectives}

In this part, we only consider input text descriptions for each example. Also, these objectives can alleviate the costs of collecting MWP corpus by constructing supervision signals without solution answers and equations. 

\paragraph{Masked Language Modeling.} We follow \citet{devlin2019bert} and introduce masked language modeling (MLM) for basic contextual representation modeling. Specially, we apply masks on 10\% of tokens, randomly replace 10\% of tokens with other tokens and keep 80\% of tokens unchanged. Later, the manipulated sentence is utilized to reconstruct the original sentence. 

\paragraph{Number Counting.} Another pre-training objective is to predict the amount of numbers that appeared in MWP description. The amount of a number corresponds with the cardinality of variable sets. This also reflects the basic understanding about the difficulty of an MWP and can act as a key contextual MWP number understanding feature. Here, we introduce a regression objective with the following formulated loss function:
\begin{equation}
\label{s1}
    L_{NumCount} = MSE(FFN(\bar{Z}), |W_q|),
\end{equation}
where $MSE$ stands for mean-squared-error and FFN stands for the feed-forward network which is made up of two fully-connected (FC) layers and one ReLU activation. We build a two-FC-layer block for each pre-training task (except MLM) and discard them during fine-tuning. $\bar{Z} \in \mathbb{R}^{h}$ is the mean vector of $Z$ and represents the encoder's overall understanding of a single MWP text description. $|W_q|$ is the number of quantities shown in the problem description $W$. 

\paragraph{Number Type Grounding.} This objective aims at linking contextual number representations with corresponding number types to tell the difference between discrete and continuous concepts/entities. For numerical reasoning in MWP solving, we only need to handle whole numbers as well as non-integer numbers (decimal, fraction and percentage). Ideas here are that whole numbers usually associate with discrete entities (for example, desks, chairs and seats) while non-integer numbers often connect with continuous concepts (for example, proportions, rate, velocity). Besides, comparisons among whole numbers got different issues compared with rational numbers. Therefore, we propose a classification objective to predict if a number is a whole number or non-integer number:
\begin{equation}
\label{s2}
    L_{NTGround} = \sum\limits_{i:W_i \in W_q}CE(FFN(Z_i), y_i),
\end{equation}
where $W_q$ contains all the numbers that appeared in $W$, and $CE$ is the cross-entropy loss for binary classification. Here, $i$ is the index when $W_i$ is a quantity, $Z_i$ is the corresponding representation vector, and $y_i$ is a binary label indicating if $W_i$ is a whole number or non-integer number.

\subsection{Weakly-supervised Objectives}
Given both text descriptions of MWPs and corresponding answers, we can model dependencies among answer number and numbers in text descriptions so that contextual representation perceive the existence of the target variable number that does not appear in the text descriptions. In detail, we design 3 novel pre-training objectives specializing in value-annotated MWPs to improve number representation in our MWP-BERT. 

\paragraph{Answer Type Prediction.} Determining the type of answer number can provide us discrete/continuous nature of target entity/concept. Thus, we want to predict type (whole/non-integer) of the answer value given global representations of an MWP (embedded in $Z$):
\begin{equation}
\label{w1}
    L_{ATPred} = CE(FFN(\bar{Z}), y_s),
\end{equation}
where $y_s$ is the ground truth label indicating the type of answer number. 

\paragraph{Context-Answer Type Comparison.} Besides the global context feature, an MWP-BERT also needs to associate context numbers and answer number (the target number does not explicitly appear in the text). Thus, another objective is proposed to predict if the quantities appeared in the MWP text fall into the same category as the answer (i.e. they are all whole or non-integer):
\begin{equation}
\label{w2}
    L_{CATComp} = \sum\limits_{i:W_i \in W_q} CE(FFN(Z_i), y_i \oplus y_s),
\end{equation}
where $\oplus$ stands for the exclusive-or operator between two binary labels to check if they are the same, the label of a quantity $y_i$ and the label of the solution value $y_s$. 

\paragraph{Number Magnitude Comparison.} Beyond type, the magnitude of a number serves as the foundation of numerical reasoning. By associating magnitudes evaluation with contextual representation, the model can get a better perception about variance over key reasoning cues like time, size, intensity and speed. Let $\dot{y_i}$ indicate if the current quantity $W_i$ is greater than the solution value or not.   
Moreover, the loss function is formulated as:
\begin{equation}
\label{w3}
    L_{NumMComp} = \sum\limits_{i:W_i \in W_q} CE(FFN(Z_i), \dot{y}_i).
\end{equation}

\subsection{Fully-supervised Objectives}
Given both equations and answers for MWPs, we can design fully-supervised training tasks to associate number representation with reasoning flows (solution equation). Mathematical equations are known to be binary tree structures with operators on root nodes and numbers on leaf nodes. The motivation is to encourage models to learn structure-aware number representations that encode the information on how to make combinations over atomic operators and numbers. We incorporate two pre-training objectives based on the solution equation tree. 

\paragraph{Operation Prediction.} The first one is a quantity-pair relation prediction task that focuses on the local feature of the equation tree. The goal is to predict the operator between two quantity nodes in the solution tree. This is in fact a classification task with 5 potential targets, i.e., $+, -, \times, \div $ and $\wedge$. The loss function of this task is:
\begin{equation}
\small
\label{f1}
    L_{OPred} = \sum\limits_{i, j} CE(FFN([Z_i;Z_j]), op(W_i, W_j)),
\end{equation}
where $i, j$ are two indexes that satisfy $W_i, W_j \in W_q$ and $[Z_i;Z_j] \in \mathbb{R}^{2h}$ is the concatenation of $Z_i$ and $Z_j$ for the quantity $W_i$ and $W_j$. $op(W_i, W_j)$ returns the operator between $W_i$ and $W_j$.

\paragraph{Tree Distance Prediction.} Another pre-training objective is to incorporate the global structure of the equation tree in a quantitative way. Inspired by \citet{DBLP:conf/naacl/HewittM19}, we consider the depth of each number and operator on the corresponding binary equation tree to be the key structure priori. Thus, we design another fully-supervised objective to utilize this information. More exactly, given the representation of two number nodes in an equation tree, this is a regression problem that predicts the distance (difference of their depth) between them. The loss is formulated as:
\begin{equation}
\small
\label{f2}
    L_{TPred} = \sum\limits_{i,j} MSE(FFN([Z_i;Z_j]), d(W_i, W_j)),
\end{equation}
where $d(W_i, W_j)$ is the distance between quantity $W_i$ and $W_j$ in the solution tree.

The final pre-training objective is the summation of Equation 2\textasciitilde8 and the masked language model.

\subsection{Fine-Tuning}
To investigate the mathematical understanding ability of our pre-training MWP-BERT, we evaluate our model of MWP solving, quantity tagging and 7 probing tasks. Moreover, we not only use BERT but also RoBERTa~\cite{liu2019roberta} as the backbone of our encoder to show the adaptiveness of proposed method.

\section{Experiments}
We present several empirical results with octopus evaluation settings~\cite{DBLP:conf/acl/BenderK20} to prove the superiority of MWP-BERT {and MWP-RoBERTa} solver. In section~\ref{experiment:mwpsolving}, we illustrate the application of both solvers in the generation scenario, MWP solving, by fine-tuning them with a specific decoder~\cite{xie2019goal}. Next, we present MWP probing tasks in section~\ref{experiment:mwpprobing} to evaluate the capability of MWP-BERT  and MWP-RoBERTa on ``understanding'' or ``capturing the meanings'' of MWPs. Finally, results and analysis about ablation study are illustrated in section~\ref{experiment:ablation}.

\paragraph{Implementation Details.}
We pre-train our model on 4 NVIDIA TESLA V100 graphic cards and fine-tune on 1 card. The model was pre-trained for 50 epochs (2 days) and fine-tuned for 80 epochs (1 day) with a batch size of 32. Adam optimizer \cite{kingma2014adam} is applied with an initial learning rate of 5e-5, which would be halved every 30 epochs. Dropout rate of 0.5 is set during training to prevent over-fitting. During testing, we use 5-beam search to get reasonable solutions. The hyper-parameters setting of our BERT and RoBERTa is 12 layers of depth, 12 heads of attention and 768 dimensions of hidden features.  For the Chinese pre-training model, we use an upgrade patch of Chinese BERT and RoBERTa which are pre-trained with the whole word masking (WWM)\footnote{\url{https://github.com/ymcui/Chinese-BERT-wwm}} \cite{cui2020revisiting}. For the English pre-training models, we use the official source on this website\footnote{\url{https://huggingface.co/bert-base-uncased} and \url{https://huggingface.co/roberta-base}}. Our code and data have been open-sourced on Github \footnote{\url{https://github.com/LZhenwen/MWP-BERT}}.

\subsection{MWP Solving}\label{experiment:mwpsolving}
\paragraph{Experiment Settings and Datasets.}
Given a textual description of a mathematical problem, which contains several known variables, MWP solving targets at getting the correct answer for the corresponding question. A solver is expected to be able to predict an equation that can exactly reach the answer value. We conduct experiment based on these benchmarks, Math23k~\cite{wang2017deep}, MathQA \cite{amini2019mathqa} and Ape-210k~\cite{zhao2020ape210k}. Since there exist many noisy examples in Ape-210k, e.g., examples without equation annotations or answer values, we re-organize Ape-210k to Ape-clean and Ape-unsolvable, where the training set of Ape-clean and the whole Ape-unsolvable are used for pre-training. For the English MWP, we use the training set of MathQA~\cite{amini2019mathqa} to perform pre-training. For the implementation of our solver, MWP-BERT is adapted as an encoder to generate intermediate MWP representation for the tree-based decoder in \citet{xie2019goal}. 

\subsection{The Ape-clean Dataset}

Ape210k is a recently released large MWPs dataset, including 210,488 problems. 
The problems in Ape210k are more diverse and difficult than those in Math23k as shown in \ref{tab:data_demo}. Not only the stronger requirement of common-sense knowledge for getting solutions, but also the missing of ground-truth solution equations or answers, will take extra obstacles for MWP solving. Among all these cases, 
the problems without answers can not be used for fully-supervised setting. Besides, the problems without annotated equations but only answer values
can be used in the weakly-supervised learning setting. 
Therefore, we follow the rules below to select the usable problems from Ape210k to construct an Ape-clean dataset, which can be used for the fully-supervised learning setting. (i). We remove all MWPs that  have no answer values nor equations.
(ii). We remove all MWPs that only have answer values without equations. 
(iii). We remove all MWPs with a problem length $m > 100$ or an answer equation length $n > 20$, as they will bring obstacles for training. 
(iv). We remove all MWPs requiring external constants except $1$ and $\pi$.
(v). We remove all duplicated problems with the MWPs in Math23k, because almost all problems in Math23k can be found in Ape-210k.
After data filtering, the \emph{Ape-clean} dataset contains 81,225 MWPs, including 79,388 training problems and 1,837 testing problems. The remaining 129,263 problems in Ape210k are regarded as \emph{Ape-unsolvable}, which can be used in the pre-training tasks in the settings of self-supervised and weakly-supervised learning.

\begin{table}
\renewcommand\arraystretch{1.05}
\centering
\begin{tabular}{|p{1.8cm}<{\centering}|p{5cm}|}
\hline
Unsolvable problem 1: & The price of a ball is 6 yuan, and the price of a basketball is less than 13 times of the ball's price. How much might the price of the basketball be? \\ 
Answer:       &  ?    \\
\hline
Unsolvable problem 2: & x is a single digit and the quotient of x72/47 is also a single digit, what is x at most? \\ 
Answer:       &  3    \\ 
\hline
Unsolvable problem 3: & In the yard there were 25 chickens and rabbits. Together they had 80 legs. How many rabbits were in the yard? \\ 
Answer:       &  (80-25)*2/(4-2) = 15    \\
\hline
\end{tabular}
\caption{This table shows three kinds of discarded MWPs in Ape210k. The first one does not have a certain answer, and the solution of the second one cannot be represented by equations. Solving the third problem requires external constants. Thus we filter those problems out in our Ape-clean dataset.}
\label{tab:data_demo}
\end{table}

\begin{table}[t]
\renewcommand\arraystretch{1.15}
\centering
\setlength{\tabcolsep}{0.03mm}{
\begin{tabular}{l|c|c|c}
\hline
            & $ \mathbf{\rm Math23k}$ & $\mathbf{\rm Math23k^*}$ & $\mathbf{\rm MathQA}$    \\
\hline
DNS         & $-$  & $58.1$  & $-$             \\
Math-EN     & $66.7$ & $-$  & $-$           \\
GTS         & $75.6$ & $74.3$ & $71.3$        \\
NS-Solver   & $75.7$ & $-$  & $-$           \\
Graph2Tree  & $77.4$ & $75.5$ & $72.0$       \\
TSN-MD      & $77.4$ & $75.1$ & $-$       \\
KA-S2T      & $76.3$ & $-$ & $-$\\
NumS2T      & $78.1$ & $-$ & $-$ \\
EEH-G2T     & $78.5$ & $-$ & $-$      \\
\hline
RPKHS     & $83.9$ & $82.2$ & $-$      \\
\hline
\textbf{Encoder pre-train}     &  &  &       \\
\quad RoBERTa  & $83.5$ & $81.7$ & $75.3$  \\
\quad BERT  & $83.8$ & $82.0$ & $75.1$\\
\quad MWP-RoBERTa  & ${84.5}$ & ${82.0}$  & $\mathbf{76.6}$\\
\quad MWP-BERT  & $\mathbf{84.7}$ & $\mathbf{82.4}$ & $76.2$  \\
\hline
\textbf{Seq2Seq pre-train}     &  &  &       \\
\quad REAL      & $82.3$ & $80.0$ & $-$      \\
\quad BERT-CL      & $83.2$ & $-$ & $76.3$      \\
\quad Gen\&Rank    & $85.4$ & $84.3$ & $-$      \\
\hline
\end{tabular}}
\caption{Comparison of answer accuracy (\%) among our proposed models and different baselines. $\mathrm{Math23k}$ column shows the results on the public test set and $\mathrm{Math23k^*}$ is 5-fold cross validation on Math23k dataset. MathQA is adapated from~\citet{li2021seeking,DBLP:journals/corr/Tan2021multilingual}. ``RoBERTa'' and ``BERT'' represent results without pre-training. ``MWP-RoBERTa'' and ``MWP-BERT'' represent first pre-training with proposed tasks and then fine-tuning.}
\label{tab:main_result}
\end{table}

\paragraph{Model Comparison.}
We first compare our approach with the most recent representative baselines on the benchmark Math23k dataset. The first baseline is \textsl{DNS} which is the pioneering work using the Seq2Seq model to solve MWPs. \textsl{Math-EN} \cite{wang2018translating} proposes an equation-normalization method and uses vanilla Seq2Seq model to get solutions. \textsl{GTS} \cite{xie2019goal} proposes a goal-driven tree-based decoder and achieves great results. \textsl{Graph2Tree} \cite{zhang2020graph} constructs two graphs during data pre-processing to extract extra relationships from text descriptions. \textsl{KA-S2T} \cite{wu2020knowledge} proposes a novel knowledge-aware model that can incorporate background knowledge. 
\textsl{NS-Solver} \cite{qin2021neural} designs several auxiliary tasks to help training. \textsl{NumS2T} \cite{wu2021math} uses explicit numerical values instead of symbol placeholder to encode quantities. \textsl{RPKHS} \cite{yu2021improving} builds hierarchical reasoning encoder in parallel with PLM encoder. \textsl{REAL} \cite{huang2021recall} proposes a human-like analogical auxiliary learning strategy. \textsl{EEH-G2T} \cite{wu2021edge} 
injects edge label information and the long-range word relationship into graph network. \textsl{Gen\&Rank} \cite{DBLP:conf/emnlp/ShenYLSJ0021} designs a multi-task learning framework for adapting BART in MWP solving. \textsl{BERT-CL} \cite{li2021seeking} 
incorporates contrastive learning strategy with PLM. To avoid the implementation error that may cause unreproducible results of baseline models, we reported the results of these baselines from the papers where they were published, as many previous papers \cite{zhang2020graph,shen2020solving} did. 

As shown in Table~\ref{tab:main_result}, our MWP-BERT achieves competitive results. It is worth noting that we perform strict pre-training paradigm in MWP solving, i.e., our results come from pre-training on the different annotated MWP examples that will be applied in further fine-tuning. Here, our pre-training only uses Ape-clean and Ape-unsolvable and fine-tuning only uses Math23k/MathQA.

\textsl{RPKHS}, \textsl{REAL} and \textsl{BERT-CL} all incorporate BERT in their model architecture, which are orthogonal to our work. Our MWP-BERT can be utilized as an MWP-specific checkpoint for their encoder part to improve their performance. Besides, \textsl{REAL}, \textsl{BERT-CL} and \textsl{Gen \& Rank} are all trying to make Seq2Seq pre-training~\cite{DBLP:conf/acl/LewisLGGMLSZ20}, which adapt both pre-training encoder as well as pre-training decoder for MWP solving. Compared with them, our model focuses on encoder pre-training and aims at obtaining better MWP representation that can be widely applied across various MWP related tasks (like quantity tagging, MWP question generation). 

Another interesting observation is that BERT-based models perform better on Chinese MWP datasets while RoBERTa-based models are good at English MWP datasets. Because the Chinese RoBERTa used in this paper is actually a BERT model that uses BERT tokenization but is trained like RoBERTa (drops the Next Sentence Prediction task). Similar behaviors can be observed in \citet{cui2020revisiting}. For English setting, RoBERTa performs better than BERT, which is consistent with conclusions raised in~\citet{liu2019roberta}.



\begin{table}[t]
\renewcommand\arraystretch{1.2}
\centering
\setlength{\tabcolsep}{1.0mm}{
\begin{tabular}{l|cc|cc}
\hline
\multicolumn{1}{c|}{\multirow{2}{*}{Model}} & \multicolumn{2}{c|}{\textbf{$\mathrm{Math23k}$}} & \multicolumn{2}{c}{Ape-clean} \\ \cline{2-5}
\multicolumn{1}{c|}{} & Equ  &Ans  & Equ  &Ans   \\ \hline
    DNS              & $50.2$ & $50.3$ & $66.2$ & $66.2$   \\
    GTS              & 70.1 & $81.4$  & $60.4$ & $73.2$   \\ 
 \hline
    RoBERTa          & $75.8$ & $88.8$ & $66.7$ &  $80.2$   \\
    BERT             & $76.7$ & $89.4$  & $67.0$ & $80.4$   \\
    MWP-RoBERTa      & $77.1$ & ${90.2}$ & $67.1$ & ${80.8}$   \\
    MWP-BERT         & $\mathbf{77.5}$ & $\mathbf{91.2}$ & $\mathbf{67.5}$ & $\mathbf{81.3}$    \\
    
\hline
\end{tabular}}
\caption{Comparison of answer accuracy (\%) between our proposed models and baselines when they are all trained by the combination of the training set from Ape-clean and Math23k dataset.}

\label{tab:result_ape}
\end{table}

\paragraph{Evaluation on Ape-clean and Math23k  when being trained by a Joint MWP Set.}

Moreover, we combine the training set of Math23k and Ape-clean to train MWP-BERT, and then measure the accuracy on the testing set of Math23k and Ape-clean separately. Results shown in Table \ref{tab:main_result} convey interesting evaluation observations. Surprisingly, 
the accuracy of our models on Math23k reaches above 90\%, which is marvelously high (previous state-of-the-art methods can hardly reach 80\% \cite{shen2020solving}). 
Compared to the results in Table \ref{tab:main_result}, even \textsl{GTS} has a higher accuracy when trained  with the big joint MWP set of Ape-clean and Math23k. 

By comparing the performance of corresponding groups between Table~\ref{tab:main_result} and Table~\ref{tab:result_ape}, we can learn that our proposed MWP-BERT pre-training paradigm can achieve more significant boosting with more training examples, which proves the effectiveness of our proposed representation learning techniques.
\begin{table*}[t]
\renewcommand\arraystretch{1.1}
\centering
\setlength{\tabcolsep}{0.6mm}{
\begin{tabular}{l|ccccccc||c}
\hline

& $\mathrm{NumCount}$ & $\mathrm{NTGround}$ & $\mathrm{ATPred}$ & $\mathrm{CATComp}$ & $\mathrm{NumMComp}$ & $\mathrm{OPred}$ & $\mathrm{TPred}$ & $\mathrm{QT}$   \\
\hline
Metric & $\mathrm{MSE}\downarrow$ & $\mathrm{Acc}\uparrow$ & $\mathrm{Acc}\uparrow$ & $\mathrm{Acc}\uparrow$ & $\mathrm{Acc}\uparrow$ & $\mathrm{Acc}\uparrow$ & $\mathrm{MSE}\downarrow$ & $\mathrm{Acc}\uparrow$   \\
\hline
BERT         & 3.08 & 0.87  & 0.75 & 0.77 & 0.77 & 0.50 & 0.97 & 84.5    \\
RoBERTa     & 3.20 & 0.86 & 0.76 & 0.78 & 0.77  & 0.51 & 0.99  &84.6     \\
\hline
MWP-RoBERTa  & 0.69 & 0.92  & 0.86 & 0.87 & 0.86 & 0.86 & 0.44 & 91.0 \\
MWP-BERT  & 0.67 & 0.92  & 0.85 & 0.87 & 0.86 & 0.87 & 0.45 & 91.5 \\
\hline
\end{tabular}}
\caption{The evaluation results on MWP-specific understanding tasks. All tasks correspond to the tasks mentioned in section~\ref{tab:result_probing}. Note that the metric for 2 tasks is mean-squared-error, while others use classification accuracy. ``QT'' stands for quantity tagging.}
\label{tab:result_probing}
\end{table*}
\subsection{Other MWP Understanding Tasks}\label{experiment:mwpprobing}
Standard MWP solving is an equation generation task. To make a sufficient validation of the effectiveness of our model on number representations learning,  MWP-specific understanding tasks are  further considered. Following ~\citet{DBLP:conf/naacl/HewittM19,DBLP:conf/emnlp/WallaceWLSG19}, we design several number probing tasks and incorporate quantity tagging~\cite{zou2019text2math} to enlarge the MWP understanding evaluation task. 

Following the motivation mentioned in section~\ref{sec:train_obj}, we re-run all the pre-training tasks as probing tasks to evaluate our modeling's understanding ability and test MWP-BERT in a zero-shot scenario, i.e. without fine-tuning the parameters of MWP-BERT and MWP-RoBERTa for the sake of fair comparison. We perform the probing evaluation on both Ape-clean and Ape-unsolvable, except that ``OPred'' and ``TPred'' are only evaluated on Ape-clean because they require equation solutions as the ground truth.
\begin{table}[H]
\renewcommand\arraystretch{1.0}
\centering
\setlength{\tabcolsep}{1.0mm}{
\begin{tabular}{c|c}
\hline
 Model           & Accuracy    \\
\hline
QT(S)         & $87.3$      \\
QT(R)         & $88.7$    \\
QT(fix)      & $87.7$    \\
QT           & $90.8$    \\
\hline
BERT & $84.5$   \\
RoBERTa  & $84.6$ \\
MWP-BERT & ${91.0}$ \\
MWP-RoBERTa  & $\mathbf{91.5}$ \\
\hline
\end{tabular}}
\caption{Comparison of tagging accuracy (\%) between our proposed models and   baselines.}
\label{tab:result_tagging}
\end{table}

Table \ref{tab:result_probing} shows the performances of 4 different PLMs on the above mentioned MWP-specific understanding tasks.
Significant improvements can be observed in all the tasks, and demonstrate the effectiveness of our proposed pre-training techniques in improving number representation of PLMs. 

Besides, we borrow an MWP-specific sequence labeling task, quantity tagging~\cite{zou2019text2math} (``QT''), to further compose MWP understanding evaluation settings. Quantity tagging \cite{DBLP:conf/acl/ZouL19} is firstly proposed to solve MWP examples with only addition and subtraction operators in their solutions. Briefly speaking, this task requires the model to assign ``+'', ``-'' or ``None'' for every quantity in the problem description and can serve as an MWP understanding evaluation tool to examine the model's understanding of each variable's logic role in the reasoning flow.  More exactly, this is also a classification task with 3 possible targets. We extract the corresponding vectors of all quantities according to their positions in encoded problem $Z$ from Equation \ref{encoder}. Next, a 2-layer feed-forward block is connected to output the final prediction. 

Following the setting in baseline method \textsl{QT} \cite{DBLP:conf/acl/ZouL19}, we perform 3-fold cross-validation and the results are given in Table \ref{tab:result_tagging}, which shows that PLMs benefits from the proposed mathematical pre-training and outperforms the baselines.

\begin{table}[t]
\renewcommand\arraystretch{1.1}
\centering
\setlength{\tabcolsep}{1.0mm}{
\begin{tabular}{p{3.7cm}|m{1.6cm}<{\centering}|m{1.6cm}<{\centering}}
\hline
    & Math23k & Ape-clean \\
\hline
Only MLM & $89.8$ & $80.1$ \\
\hline
Only self-supervised & $90.4$ & $80.9$ \\
\quad  w/o MLM & $90.1$ & $80.6$ \\
\quad w/o $NumCount$ & $89.9$ & $80.5$ \\
\quad w/o $NTGround$ & $90.1$ & $80.4$ \\
\hline
Only weakly-supervised & $90.1$ & $80.8$ \\
\quad w/o $ATPred$ & $89.7$ & $80.2$ \\
\quad w/o $CATComp$ & $89.7$ & $80.4$ \\
\quad w/o $NumMComp$ & $89.6$ & $80.5$ \\
\hline
Only fully-supervised & $91.0$ & $80.5$ \\
\quad w/o $OPred$ & $90.5$ & $80.3$ \\
\quad w/o $TPred$ & $90.6$ & $80.5$ \\
\hline
MWP-BERT & $\mathbf{91.2}$ & $\mathbf{81.3}$ \\
\hline
\end{tabular}}
\caption{The experimental results show the effectiveness of every pre-trained task. ``Only self-supervised'' means we only apply 3 tasks of self-supervised pre-training on the BERT encoder. We also investigate the influence of each task. For example, ``w/o MLM'' means only performing self-supervised pre-training and  discarding the MLM pre-training task.}
\label{tab:data_ab}
\end{table}

\subsection{Ablation Study} \label{experiment:ablation}
We run ablation study over the proposed training objectives to investigate the necessity for each of them. As Table \ref{tab:data_ab} shows, all the proposed objectives can achieve improvements individually. Moreover, only using MLM results in weaker MWP solvers on Math23k (1.4\% less) and Ape-clean (1.2\% less), which again proves the effectiveness of our proposed pre-training tasks. Since the difficulty level of MWPs is usually in proportion to their solution length, we can easily identify that a set of MWPs exhibit a long-tail distribution over solution length, as well as the  difficulty level, as shown in. Fig 3 of the Appendix. Thus,
the 87\% accuracy of human-level performance in Math23k ~\cite{wang2019template} indicates that 13\% of the MWPs are difficult to solve. Any solvers that can improve the accuracy above  87\% are making significant contribution on solving the extremely difficult cases, such as MWPs whose solutions contain $\geq 4$ variables or single variable being used multiple times.
As neural models are known to be limited at dealing with these combinational and symbolic reasoning cases~\cite{DBLP:conf/iclr/LeeSRLB20}, we exam our model on these specially difficult cases.   Due to the space limit, we attach  several   examples of these difficult cases, statistics about solution length distribution and performance for increasing length of solution equations in the Appendix.  
Besides, it is worth noting that even without MLM objective, our model is able to promote the PLM competitor. Besides, we can observe that linking equation structure and number during pre-training is certainly beneficial for solving MWPs. 

\section{Conclusion}
We propose MWP-BERT, an MWP-specific PLM model with 8 pre-training objectives to solve the number representation issue in MWP. Also, a new dataset Ape-clean is curated by filtering out unsolvable problems from Ape210k, and the filtered MWPs are useful for self- and weakly-supervised pre-training. \textcolor{black}{Experimental results show the superiority of our proposed MWP-BERT across various downstream tasks on generation and understanding. In terms of the most representative task MWP solving, our approach achieves the highest accuracy, and firstly beats human performance}. Better numerical understanding ability is also demonstrated in the probing evaluation. We believe that our study can serve as a useful pre-trained pipeline and a strong baseline in the MWP community.

\bibliography{anthology,custom}
\bibliographystyle{acl_natbib}

\clearpage
\appendix

\section{Appendix}


\begin{table}
\renewcommand\arraystretch{1.05}
\centering
\begin{tabular}{|p{2.1cm}<{\centering}|p{4.8cm}|}
\hline
Problem 1: & There are 20 questions in an exam. Solving a question correctly gets 5 points, and 1 point is deducted if the answer is wrong. Jack gets 70 points. How many questions did he get right? \\ 
Answer:       &  20-(20*5-70)/(5+1)    \\
\hline
Problem 2: & Peter is reading a book. He reads 30\% of the whole book on the first day, and 15 pages on the second day. The ratio of the number of pages that has been read to the number of pages not read is 2:3. How many pages does this book have? \\ 
Answer:       &  15/(1-((3)/(2+3))-30\%)    \\ 
\hline
Problem 3: &There are 72\% of 50 students can swim, and (3/5) of 25 girls can swim, how many percent of the boys can swim? \\
Answer:       &  (50*72\%-25*(3/5))/(50-25)    \\ 
\hline
\end{tabular}
\caption{This table shows three difficult problems in Math23k.}
\label{tab:data_difficult}
\end{table}

\subsection{Accuracy w.r.t. Solution Length }
\begin{figure}[t!]
\centering 
\includegraphics[width=0.5\textwidth]{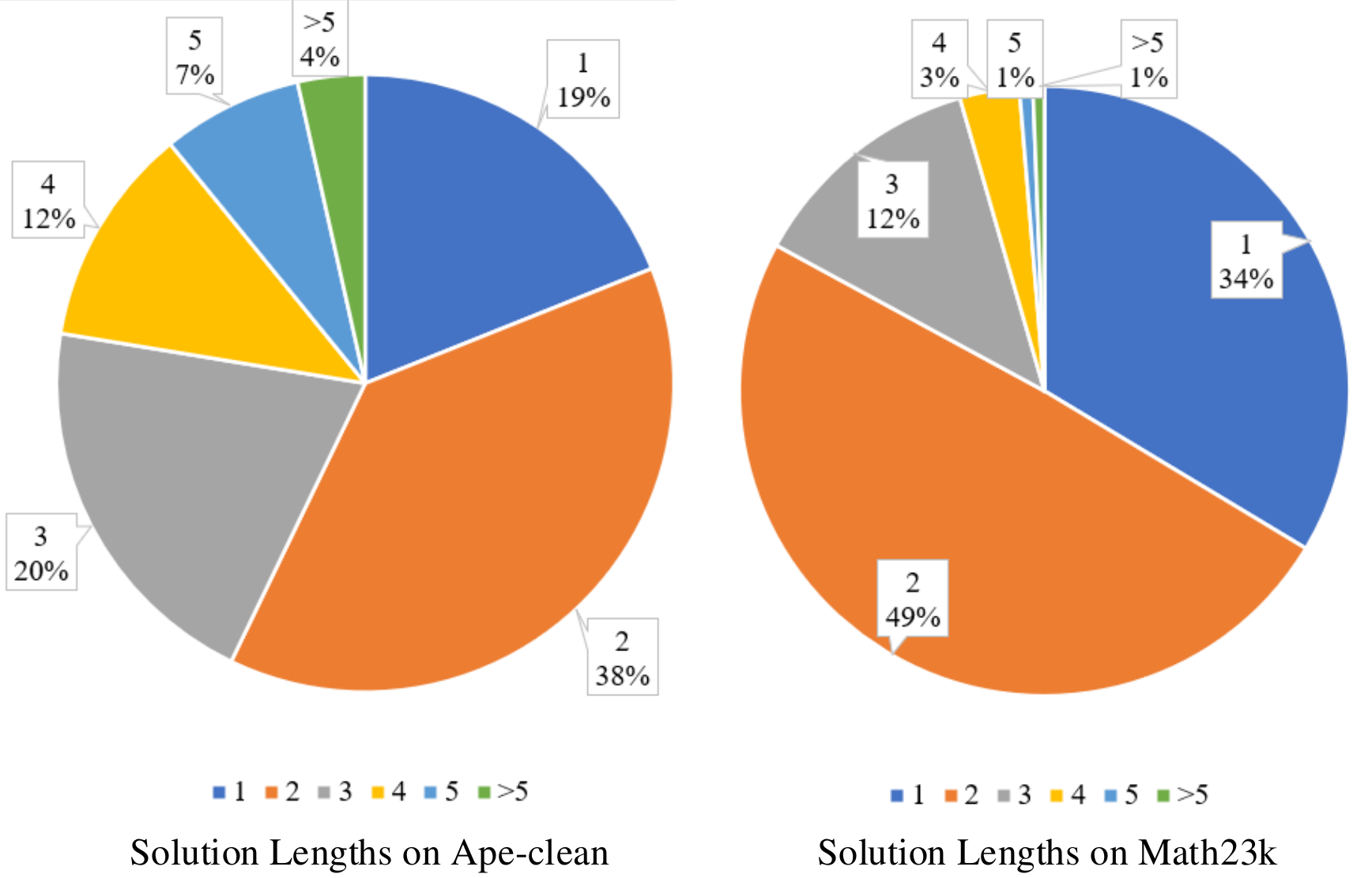}
\caption{The solution length distributions on Math23k and Ape-clean.}
\label{fig:length} 
\end{figure}

To better understand the improvement of the MWP solving performance of our model, we evaluate the problems with different lengths of solutions separately. The solution distribution details can be found in Figure \ref{fig:length} It is expected that getting longer solutions requires more comprehensive understanding and complex reasoning, like the three difficult examples   shown in Table \ref{tab:data_difficult}. The results in Table \ref{tab:data_length} demonstrate that our proposed MWP-BERT overcomes more difficult problems than the vanilla BERT model. Although the statistical improvement from BERT to MWP-BERT is marginal, our method really enhances the mathematical understanding and reasoning ability of PLMs.

\subsection{Case Study}
We perform case study as shown in Table \ref{tab:case}. Firstly, we choose a difficult problem from Math23k dataset and use 3 different solvers to solve it. Both GTS \cite{xie2019goal} and Graph2Tree \cite{zhang2020graph} fail to generate the right solution for it, while our proposed MWP-BERT solves it correctly. This example shows that our encoder has a stronger capability to understand complex MWPs to guide the tree-based decoder generate correct solutions. Secondly, when solving  a   pair of 2 similar problems (i.e., problem 1 and problem 2 in Table \ref{tab:case}),   GTS, Graph2Tree and our MWP-BERT successfully solve the former problem. However, the baseline methods GTS and Graph2Tree both fail to solve the latter one.  Our MWP-BERT generates the correct answer. This example proves that our probing tasks help the encoder to capture minor variations inside the problem description, leading to more accurate solutions.

\begin{table}
\renewcommand\arraystretch{1}
\centering
\setlength{\tabcolsep}{1.0mm}{
\begin{tabular}{|c|c|c|c|c}
\hline
 \#op & \#P  & BERT & MWP-BERT\\
\hline
0& 16 &$100$ & $100$\\
1& 331 &$95.1$ & $96.3$ \\
2& 485 &$90.7$ & $90.7$\\
3& 124  &$79.8$ & $82.3$ \\
4& 31 &$58.0$ & $67.7$\\
5& 7 &$85.7$ & $85.7$\\
>5& 6 &$33.3$ & $50$\\
\hline
\end{tabular}}
\caption{The answer accuracy of BERT and MWP-BERT on problems with different  lengths in Math23k.  \#op  denotes the number of operators in the solution. \#P is the number of problems of that kind of MWPs in the public test set of Math23k.}
\label{tab:data_length}
\end{table}

\begin{table}
\renewcommand\arraystretch{1.05}
\begin{tabular}{|p{2.1cm}<{\centering}|p{4.8cm}|}
\hline
Difficult Problem: & There are totally 48 cars and motorcycles in a parking lot. Each car has 4 wheels and each motorcycle has 3 wheels. If they have 172 wheels in total. How many motorcycles are there in the parking lot? \\
\hline
GTS:           &$x =48 + (172 - 48) / (4 - 3) $  (\xmark) \\
Graph2Tree :           &  $x = 48 - (48 - 172) / 3$ (\xmark) \\
MWP-BERT: &         $x = (48 * 4 - 172)/(4 - 3)$ (\checkmark) \\
\hline
Problem 1: & Team A and team B are working on a project together. Team A finished (4/15) of the project, and team B finished (2/15) more than Team A . How many percentage did the two teams finish in total? \\
\hline
GTS:           & $x = (4/15) + (2/15) + (4/15)$ (\checkmark) \\
Graph2Tree :           & $x = (4/15) + (2/15) + (4/15)$ (\checkmark) \\
MWP-BERT: &        $x = (4/15) + (2/15) + (4/15)$ (\checkmark) \\
\hline
Problem 2: & Team A and team B are building a road. Team A builds (4/9), and team B builds (1/9) more than team A. How many percentage does Team B build? \\
\hline
GTS:           & $x = (4/9) + (1/9) + (4/9)$ (\xmark) \\
Graph2Tree :           & $x = (4/9) + (1/9) + (4/9)$ (\xmark) \\
MWP-BERT: &        $x = (4/9)+(1/9)$ (\checkmark) \\
\hline

\end{tabular}
\caption{Our case study.}
\label{tab:case}
\end{table}


\end{document}